\documentclass[10pt,twocolumn,letterpaper]{article}

\usepackage{cvpr}
\usepackage{times}
\usepackage{graphicx, epsfig, epstopdf, pdfpages, wrapfig}
\usepackage{amsmath, amssymb}
\usepackage{color}
\usepackage{subfig}
\usepackage{algorithm,algpseudocode}
\usepackage{bm}
\usepackage{url}
\usepackage{enumitem}
\usepackage{booktabs}
\usepackage{multirow}
\usepackage{verbatim}
\usepackage{mathtools,siunitx}
\usepackage{placeins}
\usepackage{titling}

\newcommand{\cross}[1]{#1~$\times$~#1}

\usepackage[pagebackref=true,breaklinks=true,letterpaper=true,colorlinks,bookmarks=false]{hyperref}

\cvprfinalcopy 

\begin{document}

\title{Efficient Online Multi-Person 2D Pose Tracking with \\ Recurrent Spatio-Temporal Affinity Fields}

\author{
Yaadhav Raaj \qquad Haroon Idrees \qquad Gines Hidalgo \qquad Yaser Sheikh\\
The Robotics Institute, Carnegie Mellon University\\
{\tt\small \{raaj@cmu.edu, hidrees@andrew.cmu.edu, gines@cmu.edu, yaser@cs.cmu.edu\}}
}
\date{\vspace{-3ex}}

\maketitle

\begin{abstract}
We present an online approach to efficiently and simultaneously detect and track 2D poses of multiple people in a video sequence. We build upon Part Affinity Fields (PAF) representation designed for static images, and propose an architecture that can encode and predict Spatio-Temporal Affinity Fields (STAF) across a video sequence. In particular, we propose a novel temporal topology cross-linked across limbs which can consistently handle  body motions of a wide range of magnitudes. Additionally, we make the overall approach recurrent in nature, where the network ingests STAF heatmaps from previous frames and estimates those for the current frame. Our approach uses only online inference and tracking, and is currently the fastest and the most accurate bottom-up approach that is runtime-invariant to the number of people in the scene and accuracy-invariant to input frame rate of camera. Running at $\sim$30 fps on a single GPU at single scale, it achieves highly competitive results on the PoseTrack benchmarks. \footnote{\href{https://cmu-perceptual-computing-lab.github.io/spatio-temporal-affinity-fields/}{Project Page}}
\end{abstract}

\section{Introduction}

\begin{figure}
\centering
\includegraphics[width=0.475\textwidth]{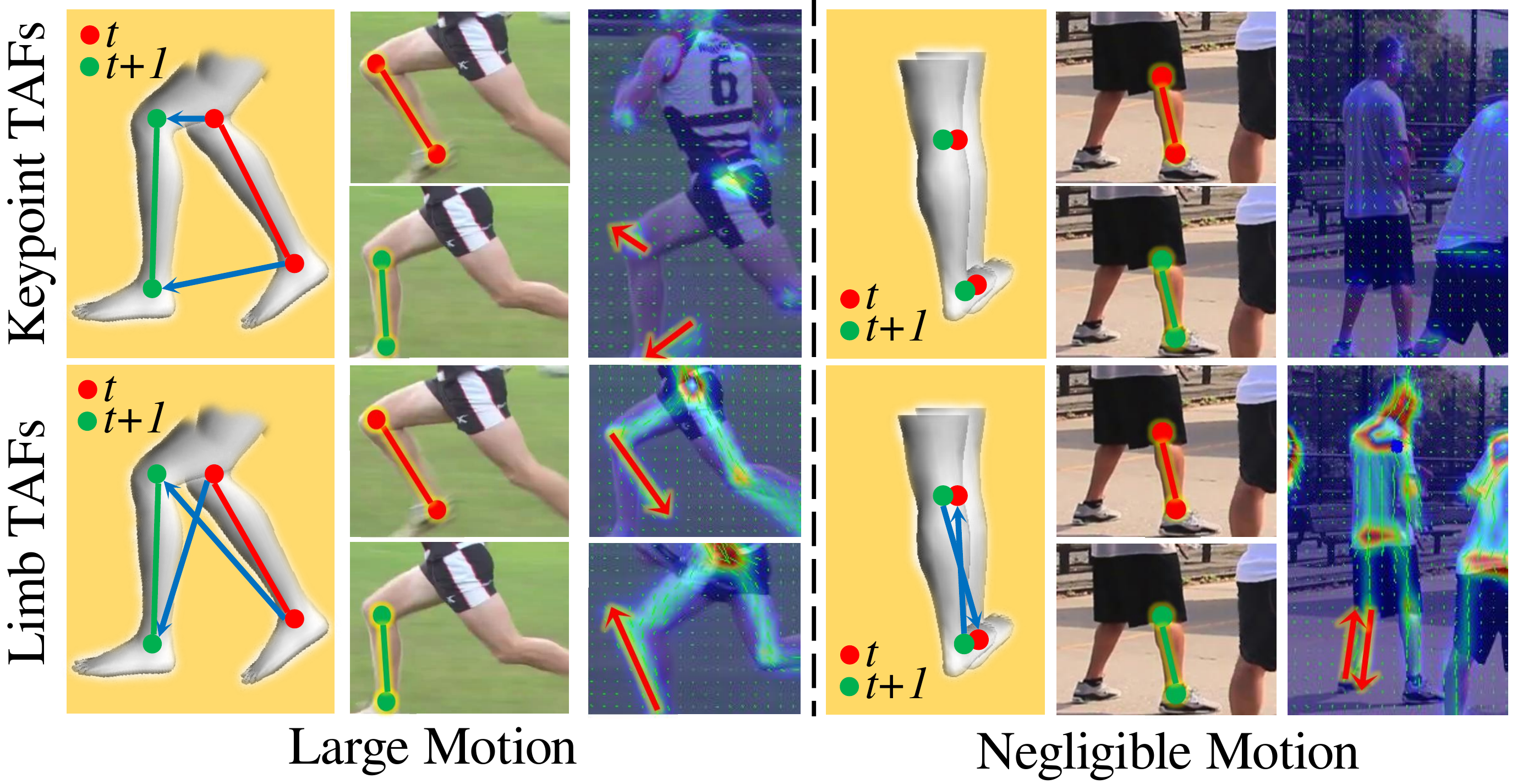}
\vspace{-.2in}
\caption{We solve multi-person human pose tracking by encoding change in position and orientation of keypoints or limbs across time as Temporal Affinity Fields (TAFs) in a recurrent fashion. \textbf{Top:} Modeling TAFs (blue arrows) through keypoints works when motion occurs but fails during limited motion making temporal association difficult. \textbf{Bottom:} Cross-linked TAFs across limbs perform consistently for all kinds of motions providing redundancy and smoother encoding for further refinement and prediction.}
\label{fig:teaser}
\end{figure}

Multi-person human pose estimation has received considerable attention in the past few years assisted by deep convolutional learning as well as COCO \cite{COCO} and MPII \cite{andriluka20142d} datasets. The recently introduced PoseTrack dataset \cite{Iqbal_CVPR2017} has provided the community with a large scale corpus of video data with multiple people in the scenes. In this paper, our aim is to utilize these towards building a truly online and real-time multi-person 2D pose estimator and tracker that is deployable and scalable while achieving high performance and requiring minimal post-processing. The potential uses include real-time and closed-loop applications with low latency where the execution is in sync with frame rate of camera such as self-driving cars and augmented reality. 

The real-time and online nature of such an approach introduces several challenges: i) scenes with multiple people demand handling of occlusion, proximity and contact as well as limb articulation, and ii) it should be runtime-invariant to the number of people in the scene. Furthermore, iii) it must be capable of handling challenges induced from video data, such as large camera motion and motion blur across frames. We build upon the Part Affinity Fields (PAFs) \cite{cao2017realtime} to overcome these challenges, which represent connections across body keypoints in static images as normalized 2D vector fields with position and orientation. In this work, we propose Temporal Affinity Fields (TAFs) which encode connections between keypoints across frames, including a unique cross-linked limb topology as seen in bottom row of Figure~\ref{fig:teaser}. In the absence of motion or when there is not enough data from previous frames, TAFs constructed between same keypoints, e.g., wrist-wrist or elbow-elbow across frames lose all associative properties (see top row of Fig.~\ref{fig:teaser}). In this case, the nullification of magnitude and orientation provides no useful information to discern between the case where a new person appears or where an existing person stops moving. This effect is compounded if these two cases occur in proximity together. However, the longer limb TAF connections allow information preservation even in the absence of motion or appearance of new people by preventing corruption of valid information with noise as the magnitude of motion becomes small. In the limiting case of zero motion, \textit{the TAF effectively collapses to a PAF}. From the perspective of a network, TAF between keypoints destroys spatial information about keypoints as motion ceases, whereas TAF across keypoints simply learns to propagate the PAF, which is a much simpler task. 

Furthermore, we work on videos in a recurrent manner to make the approach real-time, where computation of each frame leverages information from previous frames thereby reducing overall computation. Where the single-image pose estimation methods use multiple stages to refine heatmaps \cite{cao2017realtime,hourglass}, we exploit the redundant information in the video frames and divert the resources towards efficient computation of both poses and tracks across multiple frames. Thus, the multi-stage computation over images is divided over multiple frames in a video. Overall, we call this Recurrent Spatio-Temporal Affinity Fields (STAF) and it achieves highly competitive results on the PoseTrack benchmarks: [64.6\%~mAP, 58.4\%~MOTA] on single scale at $\sim$30 FPS, and [71.5\%~mAP, 61.3\%~MOTA] on multiple scales at $\sim$7 FPS on the PoseTrack 2017 validation set using one GTX 1080~Ti. As of writing, our approach currently ranks second for accuracy and at third place for tracking on the 2017 challenge \cite{posetrack-leaderboard}. Note that, our tracking approach is truly online on a per-frame basis with no post processing.

The rest of the paper is organized as follows. In Sec.~\ref{sec:RelatedWork}, we discuss related work and situate the paper in the literature. In Sec.~\ref{sec:approach}, we present details of our approach, training procedure as well as tracking and inference algorithm. Finally, we present results and ablation experiments in Sec.~\ref{sec:experiments} and conclude the paper in Sec.~\ref{sec:conclusion}.

\section{Related Work}
\label{sec:RelatedWork}

Early methods for human pose estimation localized keypoints or body parts of individuals but did not consider multiple people simultaneously \cite{andriluka2010monocular, pishchulin2013poselet,yang2013articulated,johnson2010clustered,wei2016cpm}. Hence, these methods were not adept at localizing keypoints of highly articulated or interacting people. Person detection was typically used which followed single-person keypoint detection \cite{pishchulin2012articulated,gkioxari2014using,sun2011articulated,iqbal2016multiperson}. With deep learning, human detection methods such as Mask-RCNN \cite{Detectron2018, he2017maskrcnn} were employed to directly predict multiple human bounding boxes through ROI-pooling followed by pose estimation per person \cite{Guler2018DensePose}. However, these methods suffered when people were in close proximity as bounding boxes got grouped together. Furthermore, these \textbf{top-down} methods required more computation as the number of people increased in the image, making them inadequate for real-time pose estimation and tracking.

The \textbf{bottom-up} Part Affinity Fields (PAF) method~\cite{cao2017realtime} produced a spatial encoding of pair-wise body part connections in the image space, followed by greedy bipartite graph matching for inference permitting consistent computation speed irrespective of the number of people. Person Lab~\cite{papandreou2018personlab} built upon these ideas to incorporate redundant connections on people with a less greedy inference approach getting highly competitive results on the COCO \cite{coco2014} and MPII \cite{andriluka20142d} datasets. These methods work on single images and do not incorporate any keypoint tracking or past information.

Many offline methods have been proposed to enforce temporal consistency of poses in videos \cite{Insafutdinov2017ArtTrackAM,Iqbal_CVPR2017,binxiao}. These require solving spatio-temporal graphs or incorporating data from future frames making them inadequate for online operation. Alternatively, Song \etal and Pfister \etal \cite{Pfister15a,Song:2017:ThinSlicing} demonstrate how optical flow fields could be predicted per keypoint by formulating the input to be multi-framed. LSTM Pose Machines \cite{Luo2018LSTMPose} built upon previous work demonstrating use of single stage per frame for video sequences. However, these networks did not model spatial relationship between keypoints and were evaluated on the single person Penn Action \cite{Weiyu2017} and JHMDB \cite{Jhuang:ICCV:2013} datasets. 

A different line of works explored maintaining temporal graphs in neural networks for handling multiple people \cite{Girdhar2017DetectandTrackEP,DBLP:journals/corr/abs-1805-04596}. Rohit \etal demonstrated that a 3D extension of Mask-RCNN, called person tubes, can connect people across time. However, this required applying grouped convolutions over a stack of frames reducing speed, and did not achieve better results for tracking than the Hungarian Algorithm baseline. Joint Flow \cite{DBLP:journals/corr/abs-1805-04596} used the concept of Temporal Flow Field which connected keypoints across two frames. However, it did not use a recurrent structure and explicitly required a pair of images as input increasing run-time significantly. The flow representation also suffered from ambiguity when subjects moved slowly or were stationary and required special handling of such cases during tracking. 

Top-down pose and tracking methods \cite{binxiao,wei2016cpm,chen2017cascaded,papandreou2017towards,he2017maskrcnn} have dominated the detection and tracking tasks \cite{binxiao} \cite{Xiu2018PoseFE} in PoseTrack but their speed suffered due to explicit human detection and follow-up keypoint detection for each person. Moreover, modeling long-term spatio-temporal graphs for tracking in an offline manner hurts real-time applications. None of these methods are able to report any significant runtime-to-performance measures as they cannot run in real time. In this work, we demonstrate this problem can be solved in a simple elegant single-stage network that incorporates recurrence by using the previous pose heatmaps to predict both keypoints and their spatio-temporal associations. We call this Recurrent Spatio-Temporal Affinity Fields (STAF) which not only represents the prediction of Spatial (PAF) and Temporal (TAF) Affinity Fields but also how they are refined through past information. 

\begin{figure*}[t]
\centering
\includegraphics[width=1.0\textwidth]{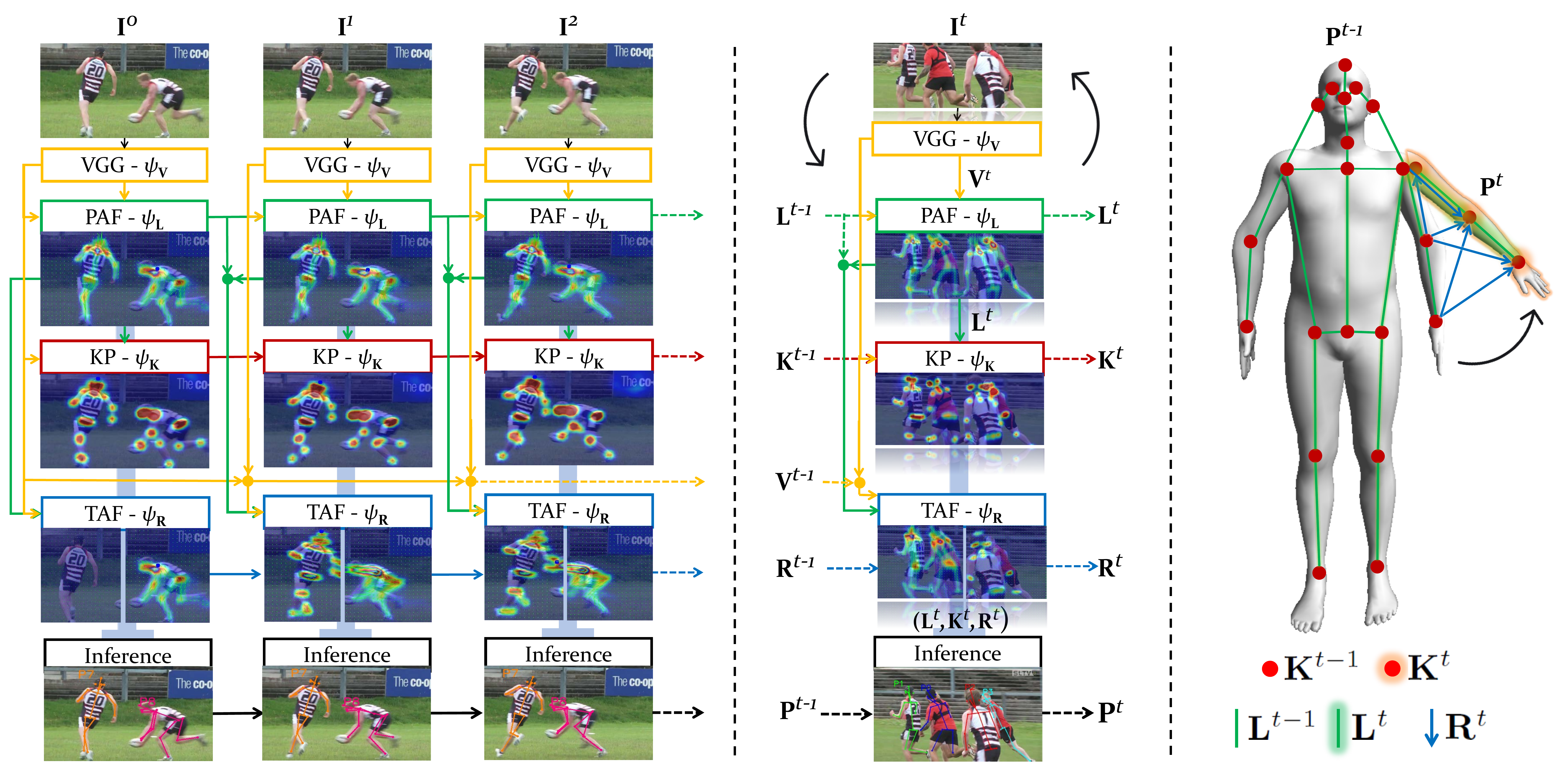}
\caption{\textbf{Left:} Training architecture for one of our models which ingests video sequences in a recurrent manner across time while generating keypoints and connections across keypoints in each frame as Part Affinity Fields (PAFs), and connections between keypoints across frames as Temporal Affinity Fields (TAFs). Together, we call this Recurrent Spatio-Temporal Affinity Fields (STAF). Each module ingests outputs from other modules in both previous and current frames (shown with arrows) and refines it. 
\textbf{Center:} During inference, our network operates on a single video frame at each time step using past information. \textbf{Right:} During inference, we use the predicted heatmaps to detect and track people. Keypoints (red) are extracted first, then associated into poses and tracklets using PAFs (green), TAFs (blue), and tracklets from previous frames.}
\label{fig:modelx}
\vspace{-.1in}
\end{figure*}

\section{Proposed Approach}
\label{sec:approach}

Our approach aims to solve the problems of keypoint estimation and tracking simultaneously in videos. We employ Recurrent Convolutional Neural Networks which we construct from four essential building blocks. Let $\mathbf{P}^t$ represent the pose of a person in a particular frame or time $t$, consisting of keypoints $\mathbf{K} = \{\mathbf{K}_1, \mathbf{K}_2, \ldots \mathbf{K}_K\}$. The Part Affinity Fields (PAFs) $\mathbf{L} = \{\mathbf{L}_1, \mathbf{L}_2, \ldots \mathbf{L}_L\}$ are synthesized from keypoints in each frame. For tracking keypoints across frames a video, we propose Temporal Affinity Fields (TAFs) given by $\mathbf{R} = \{\mathbf{R}_1, \mathbf{R}_2, \ldots \mathbf{R}_R\}$ which capture the recurrence and connect the keypoints across frames. Together, they are referred to as Spatio-Temporal Affinity Fields (STAF). These blocks are visualized in Fig.~\ref{fig:modelx} where each block is shown with a different color: the raw convolutional feature from VGG backbone \cite{simonyan2014very} are shown in amber, PAFs in green, keypoints in red and TAFs in blue.

Thus, the output of VGG backbone, PAFs, keypoints and TAFs are given by $\mathbf{V}$, $\mathbf{L}$, $\mathbf{K}$ and $\mathbf{R}$, respectively, and computed through CNNs by $\psi_\mathbf{V}$, $\psi_\mathbf{L}$, $\psi_\mathbf{K}$ and $\psi_\mathbf{R}$, respectively. The keypoint heatmaps are constructed from ground truth by placing a Gaussian kernel at the location of the annotated keypoint, whereas the PAFs and TAFs are constructed from ground truth between pairs of keypoints for each person:
\begin{equation}
\mathbf{\widetilde{L}}_{k \rightarrow k'}^{t} \coloneqq  \Omega\big(\mathbf{\widetilde{K}}_{k}^{t}, \mathbf{\widetilde{K}}_{k'}^{t}\big),\;\;
\mathbf{\widetilde{R}}_{k \rightarrow k'}^{t} \coloneqq  \Omega\big(\mathbf{\widetilde{K}}_{k}^{t-1}, \mathbf{\widetilde{K}}_{k'}^{t}\big),
\label{eq:defPAF_TAF}
\vspace{-.1in}
\end{equation}
where $\sim$ denotes the ground truth and the function $\Omega ( \cdot )$ places a directional unit vector at every pixel within a pre-defined radius of the line connecting the two keypoints.

\subsection{Video Models for Pose Estimation and Tracking} \label{subsec:videomodels}
Next, we present the three models comprising the four blocks capable of estimating keypoints and STAF. The input to each network consists of a set of consecutive frames of a video. Each block in each network consists of five \cross{7} and two \cross{1} convolution layers. Each \cross{7} layer is replaceable with the concatenation of three \cross{3} convolution layers providing the same receptive field. The first stage has a unique set of weights from subsequent frames as it cannot incorporate any previous data and also has a lower depth which was found to improve results (see Sec.~\ref{sec:experiments}). 
The VGG features are computed for each frame. For frame $\mathbf{I}^t$ at time $t$ of the video, they are computed as $\mathbf{V}^{t} = \psi_\mathbf{V}(\mathbf{I}^{t} )$.


\smallskip
\noindent\textbf{Model~I:} Given $\mathbf{V}^{t-1}$ and $\mathbf{V}^{t}$, the the following equations describe the first model:
\begin{align}
\mathbf{L}^{t} = & \;\; \psi_\mathbf{L}\big(\mathbf{V}^{t}, \psi_\mathbf{L}^{q-1}(\cdot) \big), \nonumber\\
\mathbf{K}^{t} = & \;\; \psi_\mathbf{K}\big(\mathbf{V}^{t}, \; \psi_\mathbf{L}^{q}(\cdot), \; \psi_\mathbf{K}^{q-1}(\cdot) \big), \\
\mathbf{R}^{t} = & \;\; \psi_\mathbf{R}\big(\mathbf{V}^{t-1}, \; \mathbf{V}^{t}, \; \mathbf{L}^{t-1},\; \mathbf{L}^{t},\; \mathbf{R}^{t-1} \big), \nonumber
\label{eq:model1}
\end{align}
where $\psi^q$ means $q$ recursive applications of $\psi$. In our experiments, we found that performance plateaus at $q=5$. In Model~I, PAFs are obtained by recursive application of $\psi_\mathbf{L}$ on concatenated input from VGG features and PAFs from previous stage. Similarly, keypoints depend on VGG features, keypoints from the previous stage and PAFs from the current stage. Finally, TAFs are dependent on VGG features and PAFs from both the previous and current frames, as well as TAFs from previous frame. This model produces good results but is the slowest due to recursive stages.

\smallskip
\noindent\textbf{Model~II:} Unlike Model~I with multiple applications of CNNs for PAFs and keypoints, Model~II computes the PAFs and keypoints in a single pass as visualized in Fig.~\ref{fig:modelx}:
\begin{align}
\mathbf{L}^{t} = & \;\; \psi_\mathbf{L}\big(\mathbf{V}^{t}, \mathbf{L}^{t-1} \big), \nonumber\\
\mathbf{K}^{t} = & \;\; \psi_\mathbf{K}\big(\mathbf{V}^{t}, \; \mathbf{L}^{t}, \; \mathbf{K}^{t-1} \big), \\
\mathbf{R}^{t} = & \;\; \psi_\mathbf{R}\big(\mathbf{V}^{t-1}, \; \mathbf{V}^{t}, \; \mathbf{L}^{t-1},\; \mathbf{L}^{t},\; \mathbf{R}^{t-1} \big). \nonumber
\label{eq:model2}
\end{align}

Replacing five stages with a single stage is expected to drop performance. Therefore, the multi-stage computation of PAFs and keypoints in Model~II is supplanted with output of PAFs and keypoints from the previous frames. This boosts up the speed significantly without major loss in performance as it takes advantage of the redundant information in videos, i.e., the PAFs and keypoints from previous frame are a reliable guide to the location of PAFs and keypoints in the current frame.

\smallskip
\noindent\textbf{Model~III:} Finally, the third model attempts to estimate Part and Temporal Affinity Fields through a single CNN:
\begin{align}
\mathbf{[L,R]}^{t} = & \;\; \psi_\mathbf{[L,R]}\big(\mathbf{V}^{t-1}, \mathbf{V}^{t}, \mathbf{[L,R]}^{t-1} \big), \nonumber\\
\mathbf{K}^{t} = & \;\; \psi_\mathbf{K}\big(\mathbf{V}^{t}, \; \mathbf{L}^{t}, \; \mathbf{K}^{t-1} \big),
\end{align}
where $\mathbf{[L,R]}$ implies simultaneous computation of Part and Temporal Affinity Fields through a single CNN. For Model~III, the channels corresponding to PAFs are then passed for keypoint estimation along with VGG features from current frame and keypoints from previous frame. As Model~III consists of only three blocks, it has the fastest inference, however it proved to be the most difficult to train.

\subsection{Topology of Spatio-Temporal Affinity Fields} \label{subsec:topology}

\begin{figure}
\centering
\includegraphics[width=0.4\textwidth]{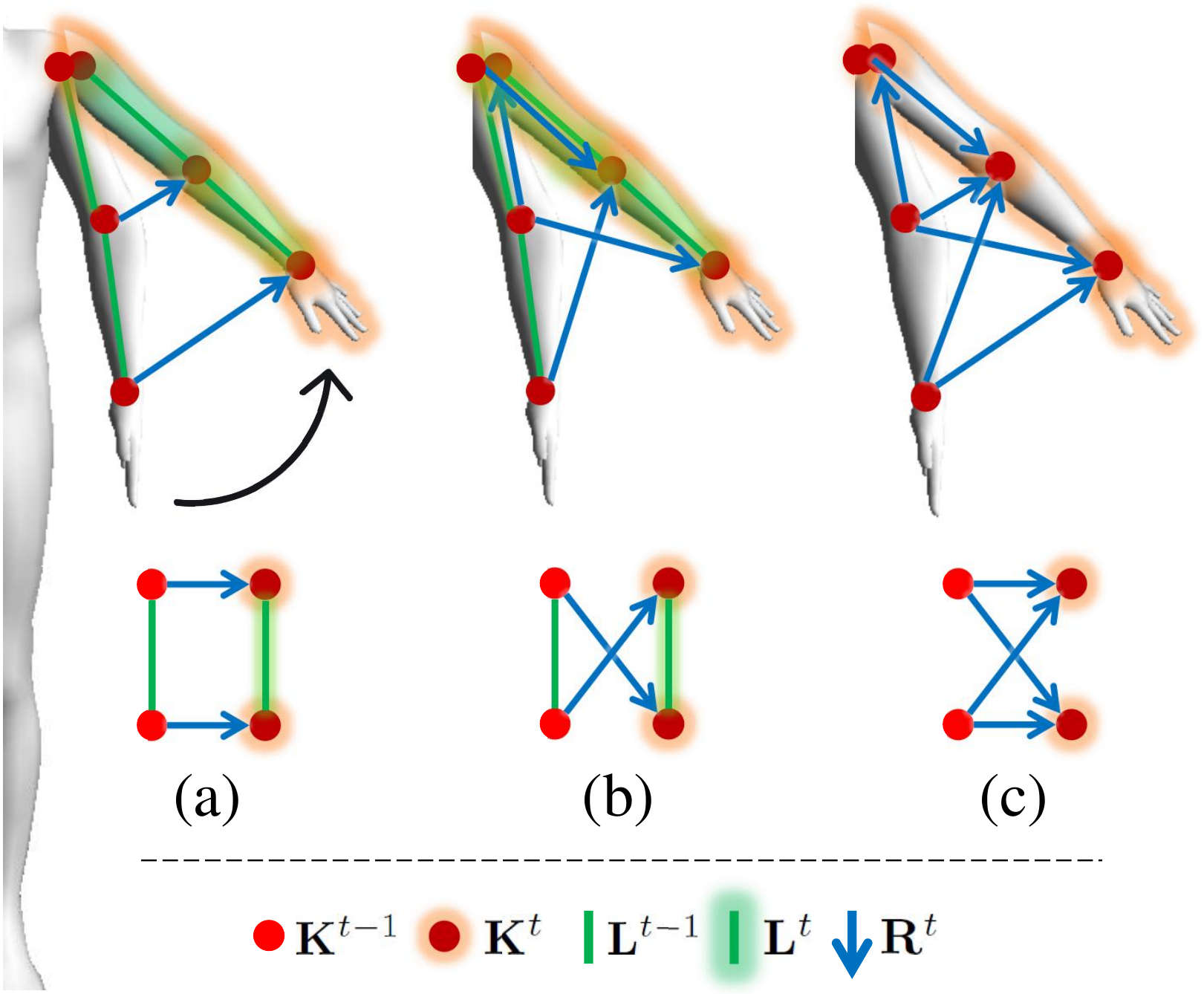}
\vspace{-.1in}
\caption{This figure illustrates the three possible topology variations for Spatio-Temporal Affinity Fields including the new cross-linked limb topology (b). 
Keypoints, PAFs and TAFs are represented by solid circles, straight lines and arrows, respectively.}
\label{fig:topologies}
\end{figure}

For our body model, we define $K=21$ body parts or keypoints which is the union of body parts in COCO and MPII pose datasets.  They include ears, nose and eyes from COCO; and head and neck from MPII.  
Next, there are several possible ways to associate and track the keypoints and STAF across frames as illustrated in Figure \ref{fig:topologies}. In this figure, solid circles represent keypoints while straight lines and arrows stand for PAFs and TAFs, respectively. Figure \ref{fig:topologies}(a) consists of TAFs between same keypoints as well as PAFs. For this topology, the number of TAFs and PAFs is 21 and 48, respectively. The TAFs capture temporal connections directly across keypoints similar to \cite{DBLP:journals/corr/abs-1805-04596}.

On the other hand, Figure \ref{fig:topologies}(b) consists of TAFs between different limbs in a cross-linked manner across frames. The number of PAFs and TAFs is 48 and 96, respectively. We also tested the topology in Figure \ref{fig:topologies}(c) which consists of 69 keypoints and limb TAFs only. This does not model any spatial links within frames across keypoints. 

\subsection{Model Training} \label{subsec:training}
During training, we unroll each model to handle multiple frames at once. Each model is first pre-trained in \textbf{Image~Mode} where we present a single image or frame at each time instant to the model. This implies multiple applications of PAF and keypoint stages to the same frame. We train with COCO, MPII and PoseTrack datasets with a batch distribution of $0.7$, $0.2$ and $0.1$, respectively, which corresponds to dataset sizes where each batch consists of images or frames from one dataset exclusively. For masking out un-annotated keypoints, we use the head bounding boxes available in MPII and PoseTrack datasets, 
and location of annotated keypoints for batches from COCO dataset. The net takes in \cross{368} images and has scaling, rotation and translation augmentations. 
Heatmaps are computed with an $\ell_2$ loss with a stride of $8$ resulting in \cross{46} dimensional heatmaps. We initialize the limb TAFs with PAFs in topology~\ref{fig:topologies}(b,c), and keypoint TAFs with zeros in topology~\ref{fig:topologies}(a,c). We train the net for a maximum of $400$k iterations.


Next, we proceed training in the \textbf{Video~Mode} where we expose the network to video sequences. For static image datasets including COCO and MPII, we augment data with video sequences that have length equal to number of times the network is unrolled by synthesizing motion with scaling, rotation and translation. We train COCO, MPII and PoseTrack in Video Mode with a batch distribution of of $0.4$, $0.1$ and $0.5$, respectively. Moreover, we also use skip-frame augmentation for video-based PoseTrack dataset where some of the randomly selected sequences skip up to $3$ frames. We lock the weights of VGG module in Video Mode. For Model~I, we only train the TAF module when training on videos. For Model~II, we train keypoint, PAF and TAF modules for $5000$ epochs, then lock all modules except TAF. In Model~III, both STAF and keypoints remain unlocked throughout the $300$k iterations. 

\subsection{Inference and Tracking} \label{subsec:inference}
The method described till now predicts heatmaps of keypoints and STAF at every frame. 
Next, we present the framework to perform pose inference and tracking across frames given the predicted heatmaps. Let the inferred poses at time $t$ 
be given by $\{\mathbf{P}^{t,1}, \mathbf{P}^{t,2}, \ldots, \mathbf{P}^{t,N}\}$ 
where the second superscript indexes over people at frame $t$. Each pose at a particular time 
consists of up to $K$ keypoints 
that become part of a pose post inference, i.e., $\mathbf{P}^{t,n} =  \{\overline{\mathbf{K}}_1^{t,n}, \overline{\mathbf{K}}_2^{t,n}, \ldots, \overline{\mathbf{K}}_K^{t,n} \}$.

The detection and tracking procedure begins with localization of keypoints at time $t$. The inferred keypoints $\overline{\mathbf{K}}^{t}$ are obtained by rescaling the heatmaps to original image resolution followed by non-maximal suppression. Then, we infer PAF weights, $\overline{\mathbf{L}}^{t}$, and those for TAF, $\overline{\mathbf{R}}^{t}$, between all pairs of keypoints in each frame defined by the given topology, i.e., 
\begin{equation}
\mathbf{\overline{L}}_{k \rightarrow k'}^{t} =
\omega\big(\mathbf{\overline{K}}_{k}^{t}, \mathbf{\overline{K}}_{k'}^{t}\big),\;\;
\mathbf{\overline{R}}_{k \rightarrow k'}^{t} =
\omega\big(\mathbf{\overline{K}}_{k}^{t-1}, \mathbf{\overline{K}}_{k'}^{t}\big),
\label{eq:infPAF_TAF}
\end{equation}
where the function $\omega(\cdot)$ samples points between the two keypoints, computes the dot product between the the mean vector of the sampled points and the directional vector from the first to the second keypoint. 

\begin{figure}[t]
\centering
\includegraphics[width=0.48\textwidth]{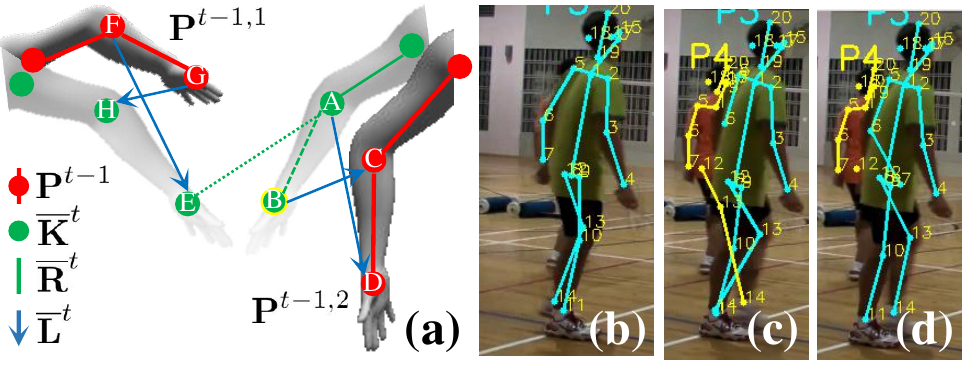}
\caption{(a) Ambiguity when selecting between two wrist locations B and E is resolved by reweighing PAFs through TAFs. (b)-(d): With transitivity, incorrect PAFs containing ankles (c) are resolved with past pose (b) resulting in (d).}
\label{fig:transitivity}
\end{figure}

Both the inferred PAFs and TAFs are sorted by their scores before inferring the complete poses 
and associating them across frames with unique ids. We perform this in a bottom-up style 
where we utilize 
poses and inferred PAFs from the previous frame 
to determine the update, addition or deletion of tracklets. Going through each PAF in the sorted list, (i) we initialize a new pose if both keypoints in the PAF are unassigned, (ii) add to existing pose if one of the keypoints is assigned, (iii) update score of PAF in pose if both are assigned to the same pose, and (iv) merge two poses if keypoints belong to different poses with opposing keypoints unassigned. Finally, we assign id to each pose in the current frame with the most frequent id of keypoints from the previous frame. 
For cases where we have ambiguous PAFs, i.e., multiple equally likely possibilities 
as seen in Figure~\ref{fig:transitivity}, we use transitivity that reweighs PAFs with TAFs to disambiguate between them, using $\alpha$ as a biasing weight. In this figure, keypoint $\{A\}$ - an elbow - is under consideration with wrists $\{B\}$ and $\{E\}$ as two possibilities. We select the strongest TAF where $\{A,B,C,D,A\}$ has a higher weight than $\{A,E,F,G,A\}$, computed as:
\begin{equation}
\overline{\mathbf{L}}_{k \rightarrow k'}^{t,n} = (1-\alpha)*\omega(\overline{\mathbf{K}}_{k}^{t-1,n},\overline{\mathbf{K}}_{k'}^{t,n})
+ \alpha*\omega(\overline{\mathbf{K}}_{k}^{t,n},\overline{\mathbf{K}}_{k'}^{t,n}). \nonumber
\label{eq:xdef55}
\end{equation} 


\section{Experiments}
\label{sec:experiments}
In this section, we present results of our experiments. Input images to networks are resized at W$\times368$ maintaining aspect ratio for single scale (SS); and W$\times736$, W$\times368$ and W$\times184$ for multiple scales (MS). The heatmaps for multiple scales are re-sized back to W$\times736$ and merged through averaging. This is followed by inference and tracking. 

\subsection{Ablation Study}
We conducted a series of ablation studies to determine the construction of our network architecture: 

\smallskip
\noindent\textbf{Filter Sizes:}  
As discussed in Sec.~\ref{sec:approach}, each block either consists of five \cross{7} layers followed by two \cross{1} layers \cite{cao2017realtime}, or each \cross{7} layer is replaced with three \cross{3} layers similar to \cite{gines} in the alternate experiment. The results are shown in Table~\ref{table:3x3vs7x7}. We run single frame inference on Model~I and find the \cross{3} filter size to be $2\%$ more accurate than \cross{7}, with significant boosts in average precision of knee and ankle keypoints. It is also $40\%$ faster while requiring $40\%$ more memory.

\begin{table}[h]
\begin{center}
\resizebox{0.475\textwidth}{!}{
\setlength{\tabcolsep}{4pt}
\begin{tabular}{l | c c c c c c c | c | c}
\hline
\textbf{Method} & \textbf{Hea} & \textbf{Sho} & \textbf{Elb} & \textbf{Wri} & \textbf{Hip} & \textbf{Kne} & \textbf{Ank} & \textbf{mAP} & \textbf{fps} \\
\hline
Model~I - 3x3 & 75.7 &73.9 &67.8 &56.3 &66.8 &62.3 &56.9 &66.3 &14\\
\hline
Model~I - 7x7 & 76.0 &73.3 &66.4 &54.0 &63.4 &59.2 &52.2 &64.3 &10\\
\hline
\end{tabular}}
\end{center}
\vspace{-.2in}
\caption{This table shows results for experiments with the two filter sizes on PoseTrack 2017 validation set.}
\label{table:3x3vs7x7}
\end{table}


\smallskip
\noindent\textbf{Video Mode / Depth of First Stage:} Next, we report results when training in \textit{Image Mode (Im)} using single images, and when we continue training beyond images while exposing the network to videos and augmenting with synthetic motion in the \textit{Video Mode (Vid)}. During testing, the network is run recurrently on video sequences with one frame per stage. Model~II is deployed for these experiments. We find that by exposing the network to video sequences for $5000$ iterations, we were able to boost the mAP as seen in Table~\ref{table:imvsvid} and Fig.~\ref{fig:ghosting}. We also find that if we use the same depth, i.e., number of channels for the first frame as the other frames (128-128), the network was not able to generalize well to recurrent execution ($56.6$ mAP) when trained with Image Mode. When reducing the depth for the first frame to one-half, i.e. (64-128), we found that the generalization to videos was better ($62.6$ mAP). When trained with Video Mode, mAP increased further to $64.1$. We reason that the 64-depth modules produced relatively vague outputs which gave sufficient room for the subsequent modules in the following frames to process and refine the heatmaps yielding a boost in performance. Furthermore, this also highlights the importance of incorporating shot change detection and running the first stage at each shot change. 

\begin{table}[h]
\begin{center}
\resizebox{0.475\textwidth}{!}{
\setlength{\tabcolsep}{2.5pt}
\begin{tabular}{l | c c c c c c c | c | c}
\hline
\textbf{Method} & \textbf{Hea} & \textbf{Sho} & \textbf{Elb} & \textbf{Wri} & \textbf{Hip} & \textbf{Kne} & \textbf{Ank} & \textbf{mAP} & \textbf{fps} \\
\hline
Im - 7x7 - 128-128 & 74.6 &69.6 &55.5 &40.2 &56.4 &47.2 &44.0 &56.6 &27\\
\hline
Vid - 7x7 - 128-128 & 76.2 &71.6 &64.5 &51.9 &62.6 &59.3 &52.5 &63.6 &27\\
\hline
Im - 7x7 - 64-128 & 73.5 &72.2 &63.8 &52.1 &62.7 &57.3 &51.1 &62.6 &27\\
\hline
Vid - 7x7 -  64-128 & 75.8 &73.4 &65.5 &53.8 &64.2 &58.4 &51.4 &64.1 &27\\
\hline
Im - 3x3 -  64-128 & 73.5 &72.5 &65.0 &52.7 &63.7 &57.7 &53.2 &63.4 &35\\
\hline
Vid - 3x3 -  64-128 & 75.4 &73.2 &67.4 &55.0 &63.9 &58.4 &53.5 &64.6 &35\\
\hline
\end{tabular}}
\end{center}
\vspace{-.2in}
\caption{This table shows single-scale performance using Model~II before and after training with videos, filter sizes, as well as different depths for first stage.}
\vspace{-5pt}
\label{table:imvsvid}
\end{table}

\begin{figure}[t]
\centering
\includegraphics[width=0.45\textwidth]{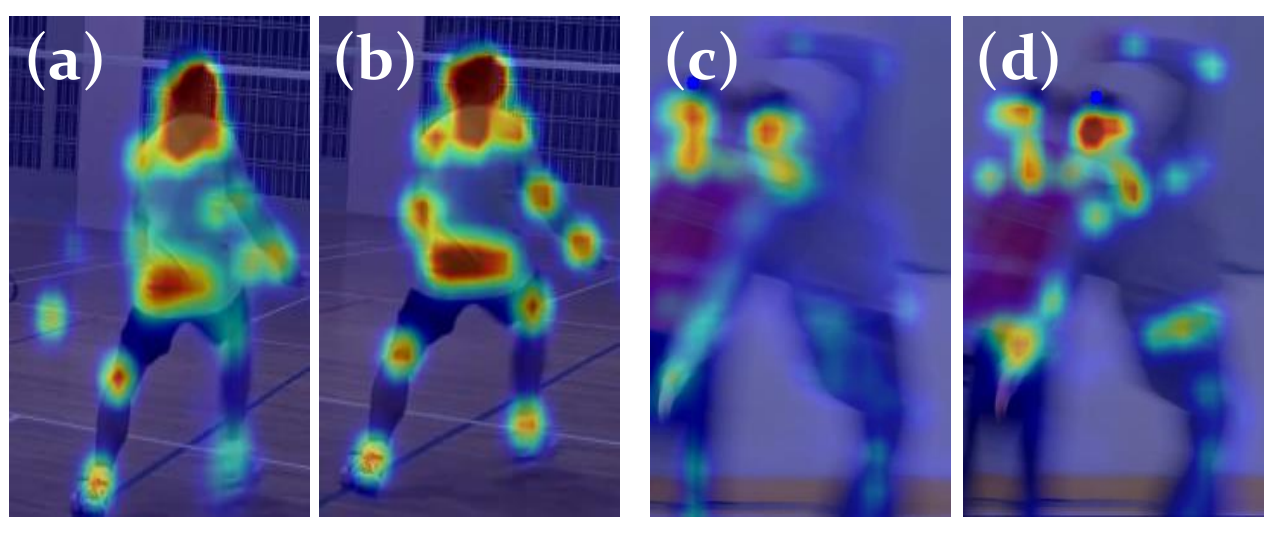}
\vspace{-.1in}
\caption{Improvement in quality of heatmaps before (a,c) and after (b,d) the network is exposed to videos and synthetic motion augmentation. We observe better peaks and less noise across both PAF and keypoint heatmaps.}
\label{fig:ghosting}
\end{figure}
 
\begin{figure}[t]
\centering
\vspace{-.1in}
\includegraphics[width=0.475\textwidth]{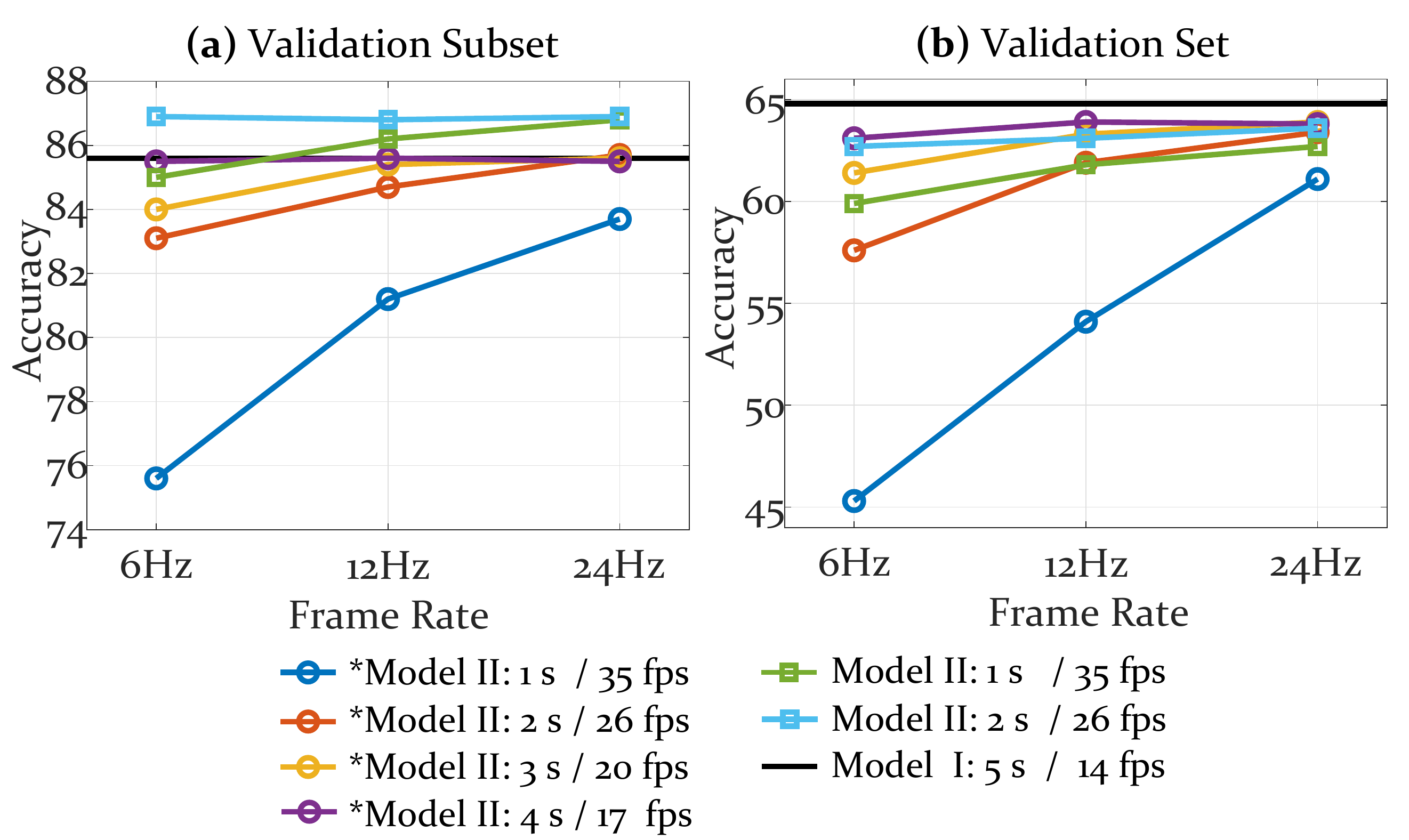}
\caption{These graphs show mAP curves as a function of frame rates of camera, i.e., the rate at which an original 24Hz video is input to the method. The flat black line shows the performance of five-stage Model~I, while `*' in the legend indicates training using Image Mode only.}%
\label{fig:cameraframerate}
\end{figure}

\smallskip
\noindent\textbf{Effect of Camera Frame Rate on mAP:} For these experiments, we studied how the frame rate of the camera and number of stages affect the accuracy of pose estimation.  
With a high frame rate, the apparent motion between frames is smooth, which becomes relatively abrupt at low frame-rates. Therefore, the heatmaps from previous frames would not be as useful at low frame-rates. We tested this hypothesis with Model~I (five stages of the same modules without ingesting previous frame heatmaps), and Model~II (different number of stages with each ingesting heatmaps from previous frame). We also evaluate the influence of training with Image and Video modes in Figure~\ref{fig:cameraframerate}.

Fig.~\ref{fig:cameraframerate}(a) shows results on a subset of ten sequences where the human subjects comprised at least 30\% of the frame height in the PoseTrack 2017 validation set. Fig.~\ref{fig:cameraframerate}(b) presents results on the entire validation set. The original videos were assumed to run at the film-standard 24 Hz, hence we ran experiments by varying frame rates at 24, 12 and 6 Hz through sub-sampling. The ground truth has been annotated at 6 Hz. As expected, accuracy is proportional to video frame rate and number of stages. 
When the Model~II was trained in Image Mode, we observed small increments in accuracy until at four stages, it peaks at the same level as Model~I. Upon training with Video Mode, it surpasses this accuracy peaking earlier at two stages.

When considering the entire validation set, the approach is still able to reap the benefits of more stages and training in Video Mode as can be seen in Fig.~\ref{fig:cameraframerate}(b). However, it was barely able to reach the accuracy of the much slower Model~I. For the validation set, the accuracy reduced when including sequences with smaller apparent size of humans. These sequences usually were more crowded as well, and passing in the previous heatmaps seemed to hurt the performance. The body parts of small-sized humans only occupied a few pixels in the heatmaps and the normalized direction vectors were inconsistent and random across frames. 

\smallskip
\noindent\textbf{Influence of Topology / Model Type in Tracking:} Next, we report experiments on different combinations of topology defined in Fig.~\ref{fig:topologies} with the three models presented in Sec.~\ref{subsec:videomodels}, both for pose estimation and tracking evaluated using mean Average Precision (mAP) and Multiple Object Tracking Accuracy (MOTA) metrics 
in Table~\ref{table:model_topology}. 
We found an improvement in tracking using limb TAFs in Topology B versus keypoint TAFs in Topology A. As highlighted in Fig.~\ref{fig:teaser}, Topology A lacks associative properties when a keypoint has minimal motion or when a new person appears. Although we enforced spatial constraint that joint locations should be close in consecutive frames, and adjusted it according to scale (similar to \cite{DBLP:journals/corr/abs-1805-04596}), this still resulted in false positives since it is difficult to disambiguate between a newly detected person and some nearby stationary person. Furthermore, where motion of a person tended to be small, Topology A resulted in jittery and noisy vectors causing more reliance on pixel distances. This was further exacerbated by recurrence where accumulation of noisy vectors from previous frame heatmaps deteriorated associative ability of Temporal Affinity Fields. Table~\ref{table:model_topology} also shows results for Topology C which significantly under-performed compared to Topology B. Since it exclusively consists of limb and joint TAFs without any spatial components, this makes keypoint localization and association rather difficult. 

Topology B solves all of these problems elegantly. The longer cross-linked limb TAF connections preserve information even in the absence of motion or appearance of new people since the TAF effectively collapses to a PAF in such cases. This allows us to avoid association heuristics and makes the problem of new person identification trivial. 
With this representation, recurrence was observably beneficial due to true and consistent representation irrespective of magnitude of motion. As a side-advantage, this also allowed us to warm-start the TAF input with PAF providing more reliable initialization for tracking in the first frame.


For Model~III, training beyond 5000 iterations gradually begins to harm the accuracy of the pose estimation resulting in reduced tracking performance as well. This is primarily due to the disparity in the amount of diverse data between COCO / MPII and PoseTrack datasets. For Model~II, if we train on keypoints and PAFs modules and lock their weights afterwards, then follow with training only the TAF, this results in better performance with a significant boost in speed as well. Although Model~I outperformed the other models with five stages for keypoints and PAFs; and a single recurrent stage for TAFs, however this comes at the expense of speed. Furthermore, we observe that an increase in mAP ends up sub-linearly increasing the MOTA as well.

\begin{table}[h]
\begin{center}
\resizebox{0.475\textwidth}{!}{
\setlength{\tabcolsep}{6pt}
\begin{tabular}{l | c c |c |c |c}
\hline
\textbf{Method} & \textbf{Wrist-AP} & \textbf{Ankles-AP} & \textbf{mAP} & \textbf{MOTA} & \textbf{fps} \\
\hline
Model~I-A &56.2 &56.4 &66.0 &58.5 &14\\
Model~I-B &56.3 &56.9 &66.3 &59.4 &13\\
Model~II-A &54.9 &53.0 &64.4 &57.4 &28\\
Model~II-B &55.0 &53.5 &64.6 &58.4 &27\\
Model~III-B &51.9 &49.5 &61.6 &57.8 &30\\
Model~III-C &42.5 &40.5 &55.2 &49.9 &36\\
\hline
\end{tabular}}
\end{center}
\vspace{-.2in}
\caption{This table shows pose estimation and tracking performance for combinations of model types and topologies.}
\vspace{-.1in}
\label{table:model_topology}
\end{table}

\smallskip
\noindent\textbf{Effect of Video Rate and Number of People on Tracking:} Finally, we performed a study on how the frame rate of the camera affects tracking accuracy, since a lower frame rate would require longer associations in pixel space.

\begin{figure}
\centering
\includegraphics[width=0.475\textwidth]{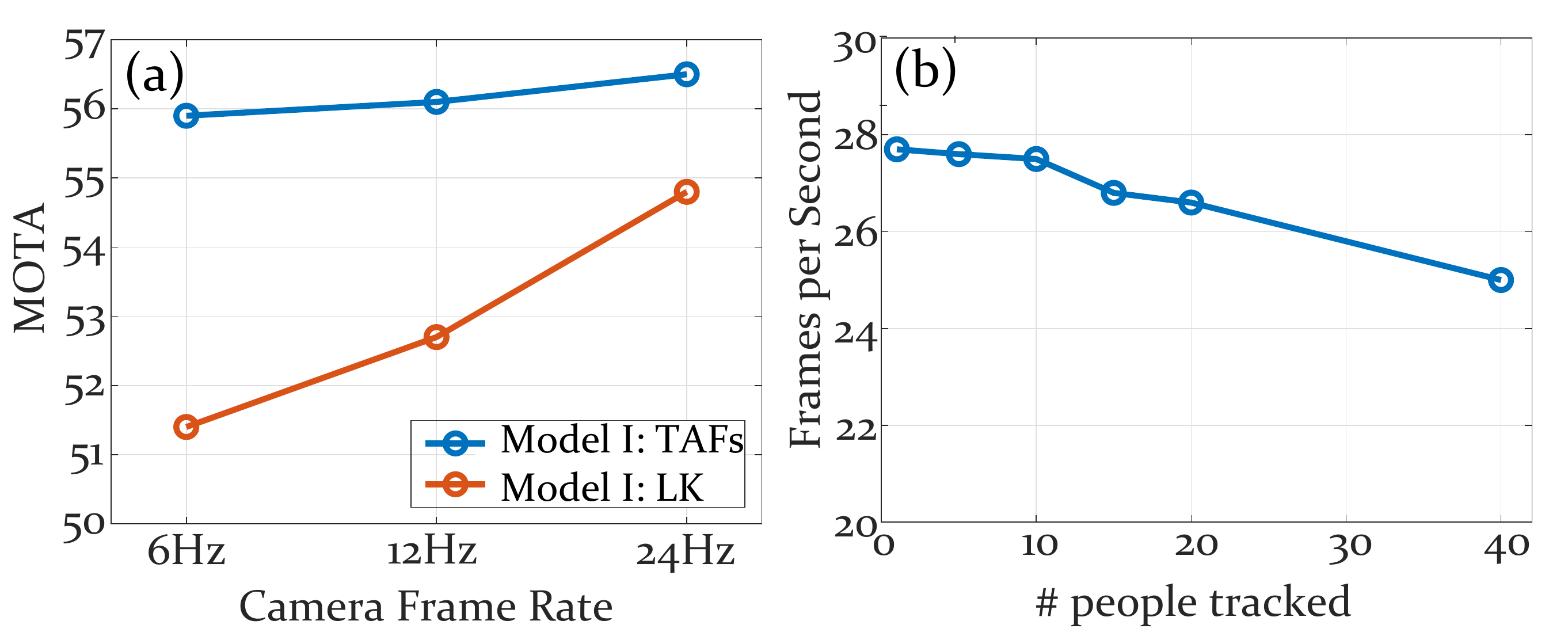}
\caption{(a) This graph shows MOTA as a function of video frame rate for Temporal Affinity Fields (TAFs) and Lukas-Kanade (LK) tracker. The performance of TAFs is virtually invariant to frame rate or alternatively to the amount of motion between frames. (b) Our approach is effectively runtime-invariant to the number of people in the scene.}
\label{fig:LK}
\vspace{-.1in}
\end{figure}

We ran Lukas Kanade (LK) as a baseline tracker by replacing the TAF Module in Model~I with LK (\cross{21} window size; $3$ pyramid levels). Initially, we observe that there is roughly $2.0\%$ improvement in MOTA as seen in Fig.~\ref{fig:LK}(a). 
However, we note that 
around 20\% of the sequences have significant articulation and camera movement, where TAFs outperformed LK as the latter was not able to match keypoints across large displacements whereas TAFs found matches due to stronger descriptive power. TAFs were able to maintain tracking accuracy even with low frame-rate cameras, but with LK the MOTA drops off significantly (see Fig.~\ref{fig:LK}(a)). Furthermore, Fig.~\ref{fig:LK}(b) suggests that our approach is nearly runtime-invariant to number of people in the frame making it suitable for crowded scenes.

\subsection{Comparison}
\begin{figure*}[t]
\centering
\includegraphics[width=.92\textwidth]{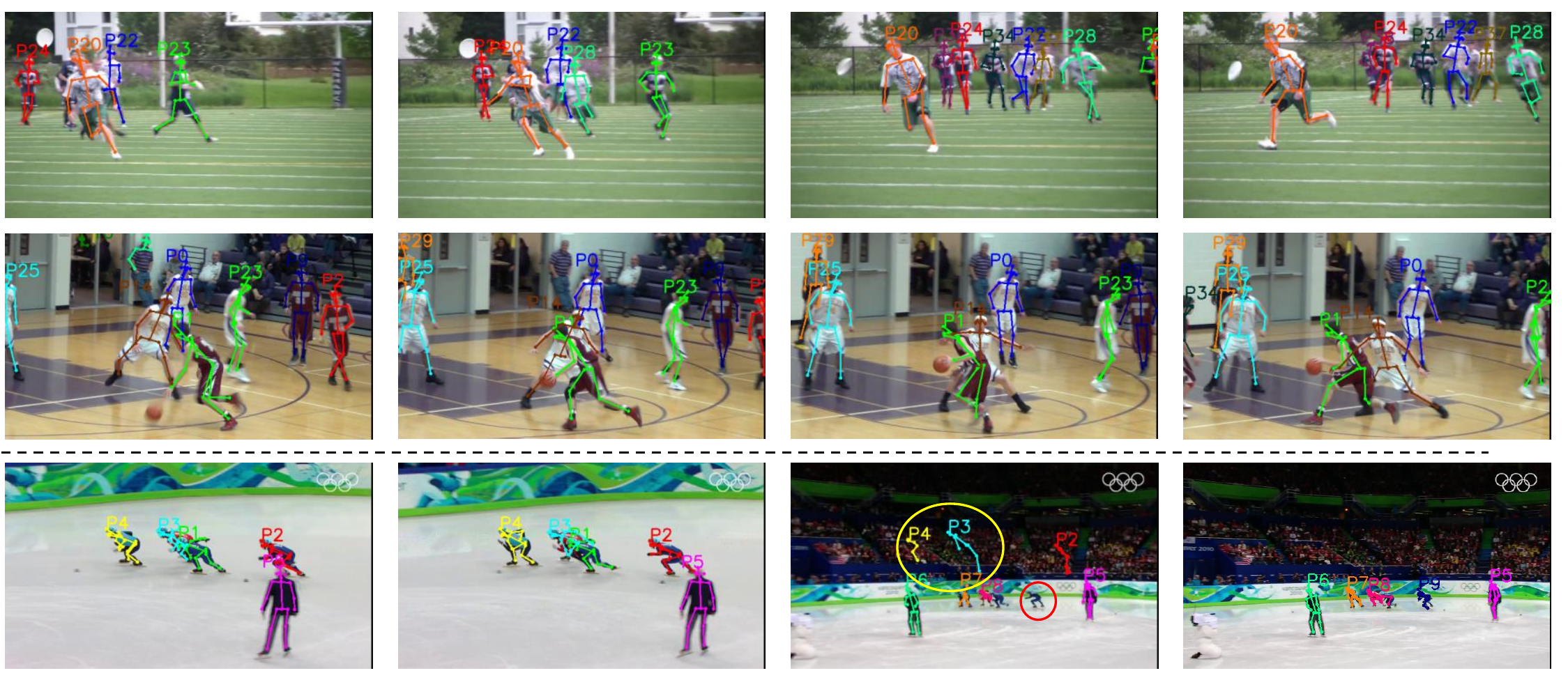}
\vspace{-.1in}
\caption{Three example cases of tracking at $\sim$30 FPS on multiple targets. \textbf{Top / Middle:} Observe that tracking continues to function despite large motion displacements and occlusions. \textbf{Bottom:} A failure case where abrupt scene change causes ghosting, where previously tracked person appears in the new frame. This issue can be rectified through a warm-start.} 
\label{fig:qual}
\end{figure*}

We present results on PoseTrack dataset in Table~\ref{table:PoseTrackComparison} for 2017 validation set (top), 2017 test set (middle) and 2018 validation set (bottom). FlowTrack, JointFlow and PoseFlow are included as comparison in this table. FlowTrack is a top-down approach which means human detection is performed first followed by pose estimation. Due to this reason, it is significantly slower than bottom-up approaches such as ours. Model~{II-B} with single scale is competitive with other bottom-up approaches while being 270\% faster. However, multi-scale (MS) processing boosts performance by $\sim$6\% and $\sim$1.5\% for mAP and MOTA, respectively. We are also able to achieve competitive results on the PoseTrack 2018 Validation set while maintaining the best speeds amongst all reported results. Note that PoseTrack 2018 Test set was not released to public at the time of submission of this paper. Figure~\ref{fig:qual} shows some qualitative results.


\section{Conclusion}
\label{sec:conclusion}

In this paper, we first motivated recurrent Spatio-Temporal Affinity Fields (STAF) as the right approach for detection and tracking of articulated human pose in videos, especially for real-time reactive systems. We showed that leveraging the previous frame data within a recurrent structure and training on video sequences yields as good results as a multi-stage network albeit at much lower computation cost. We also demonstrated the stability of tracking accuracy at reduced frame rates for the TAF formulation, due to its ability to correlate keypoints over large pixel distances. This implies that our method can be deployed on low-power embedded systems which may not be able to run large networks at high frame rates, yet are able to maintain reasonable accuracy. Our new cross-linked limb temporal topology is able to generalize better than previous approaches due to strong associative power with PAF being a special case of TAF. We are also able to operate at the same consistent speed irrespective of the number of people due to bottom-up formulation. 
For future work, we plan to embed a re-identification module to handle cases of people leaving and reappearing in a camera view. Furthermore, detecting and triggering warm-start at every shot change has the potential to boost pose estimation and tracking performance. 

\begin{table}
\begin{center}
\resizebox{0.475\textwidth}{!}{
\setlength{\tabcolsep}{4pt}
\begin{tabular}{c l | c c |c |c |c}
\cline{2-7}
& \textbf{Method} & \textbf{Wrist-AP} & \textbf{Ankles-AP} & \textbf{mAP} & \textbf{MOTA} & \textbf{fps} \\
\cline{2-7}
\multirow{5}{*}{\rotatebox[origin=c]{90}{~Top-Down}}&\multicolumn{5}{ c }{PoseTrack 2017 Validation} \\
&Detect-and-track \cite{Girdhar2017DetectandTrackEP} &51.7 &49.8 &60.6 &55.2 &1.2\\
&FlowTrack - 152 \cite{binxiao} & 72.4 &67.1 &76.7 &65.4 &-\\
&FlowTrack - 50 \cite{binxiao} & 66.0 &61.7 &72.4 &62.9 &-\\
&MDPN - 152 \cite{mdpn2018} &77.5 &71.4 &80.7 &66.0 &-\\
\cline{2-7}
\multirow{6}{*}{\rotatebox[origin=c]{90}{~Bottom-Up}}
&PoseFlow \cite{Xiu2018PoseFE} &61.1 &61.3 &66.5 &58.3 &10*\\
&JointFlow \cite{DBLP:journals/corr/abs-1805-04596} &- &- &69.3 &59.8 &0.2\\
&Model~II-B (SS) &55.0 &53.5 &64.6 &58.4 &27\\
&Model~I-B (SS) &56.8 &56.8 &66.3 &59.4 &13\\
&\textbf{Model~II-B (MS)} &\textbf{62.9} &\textbf{60.9} &\textbf{71.5} &\textbf{61.3} &\textbf{7}\\
&\textbf{Model~I-B (MS)} &\textbf{65.0} &\textbf{62.7} &\textbf{72.6} &\textbf{62.7} &\textbf{2}\\
\cline{2-7}
\multirow{4}{*}{\rotatebox[origin=c]{90}{~Top-Down}} &\multicolumn{5}{ c }{PoseTrack 2017 Testing} \\
&Detect-and-track \cite{Girdhar2017DetectandTrackEP} &- &- &59.6 &51.8 &1.2\\
&Flowtrack - 152 \cite{binxiao} &70.7 &64.9 &73.9 &57.6 &-\\
&Flowtrack - 50 \cite{binxiao} &65.1 &60.3 &70.0 &56.4 &-\\
\cline{2-7}
\multirow{6}{*}{\rotatebox[origin=c]{90}{~Bottom-Up}}&PoseTrack \cite{baseline2018} &54.3 &49.2 &59.4 &48.4 &-\\
&BUTD \cite{butd} &52.9 &42.6 &59.1 &50.6 &-\\
&PoseFlow \cite{Xiu2018PoseFE} &59.0 &57.9 &63.0 &51.0 &10*\\
&JointFlow \cite{DBLP:journals/corr/abs-1805-04596} &53.1 &50.4 &63.3 &53.1 &0.2\\
&\textbf{Model~II-B (MS)} &\textbf{62.8} &\textbf{59.5} &\textbf{69.6} &\textbf{52.4} &\textbf{7}\\
&\textbf{Model~I-B (MS)} &\textbf{65.0} &\textbf{60.7} &\textbf{70.3} &\textbf{53.8} &\textbf{2}\\
\cline{2-7}
\multirow{5}{*}{\rotatebox[origin=c]{90}{~Bottom-Up}}&\multicolumn{5}{ c }{PoseTrack 2018 Validation} \\
&Model~II-B (SS) &56.2 &54.2 &63.7 &58.4 &27\\
&Model~I-B (SS) &58.3 &56.7 &64.9 &59.6 &13\\
&Model~II-B (MS) &62.7 &60.6 &69.9 &59.8 &7\\
&Model~I-B (MS) &64.7 &62.0 &70.4 &60.9 &3\\
\cline{2-7}
\end{tabular}}
\end{center}
\vspace{-.1in}
\caption{This table shows comparison on the PoseTrack datasets. For our approach, we report results with Models I / II and Top. B. The last column shows the speed in frames per second (* excludes pose inference time). FlowTrack is a \textbf{top-down} approach using ResNet-152 (or 50); whereas JointFlow, PoseFlow and our approach are \textbf{bottom-up}.}
\vspace{-.2in}
\label{table:PoseTrackComparison}
\end{table}

\smallskip\noindent\textbf{Acknowledgment:} Supported by the Intelligence Advanced Research Projects Activity (IARPA) via Department of Interior/ Interior Business Center (DOI/IBC) contract number D17PC00340. The U.S. Government is authorized to reproduce and distribute reprints for Governmental purposes notwithstanding any copyright annotation/herein.
Disclaimer: The views and conclusions contained herein are those of the authors and should not be interpreted as necessarily representing the official policies or endorsements, either expressed or implied, of IARPA, DOI/IBC, or the U.S. Government.



\title{(Supplementary Material)}

\maketitle


\section{Model Training}

During training, we unroll each model so that it can handle multiple frames at once. Each model is first pre-trained in \textbf{Image Mode} where we present a single image or frame at each time instant to the model. This implies multiple applications of PAF/KP stages to the same frame. We train with COCO, MPII and PoseTrack datasets with a batch distribution of $0.7$, $0.2$ and $0.1$, respectively, matching dataset sizes, where each batch consists of images or frames from one dataset exclusively. We use the head bounding box information in MPII and Posetrack datasets to mask out the eyes/nose/ears in the background heatmap channel when considering the MPII and PoseTrack batches, and mask out the neck and top head positions using the annotated eyes/nose/ears keypoints for COCO batches. The net is input images of \cross{368} dimensions and has scaling, rotation and translation augmentations, where regions not annotated are masked out. Heatmaps are computed with an $\ell_2$ loss with a stride of $8$ resulting in \cross{46} dimensional heatmaps. In topology (b) and (c), we initialize the TAF with PAF, and zeros for (a). We train the net for max of $400$k iterations.


\begin{table}[h]
\begin{center}
\resizebox{0.475\textwidth}{!}{
\setlength{\tabcolsep}{4pt}
\begin{tabular}{l | c c }
\hline
\textbf{Param} & \textbf{Image Mode} & \textbf{Video Mode} \\
\hline
Input Resolution &368x368 &368x368 \\
\hline
Heatmap Resolution &46x46 &46x46 \\
\hline
Data Dist. (COCO, MPII, PT) & 0.7,0.2,0.1 &0.4,0.1,0.5 \\
\hline
Frame Skip Augmentation &- &3 \\
\hline
Scaling &[$0.7x, 1.3x$] &[$0.7x, 1.3x$] \\
\hline
Rotation &[$\ang{-30},\ang{30}$] &[$\ang{-20},\ang{30}$] \\
\hline
Translation &[$-30,30$] &[$-50,50$] \\
\hline
VGG Learning Rate &0.00004 &0.0 \\
\hline
PAF/TAF/KP Learning Rate &0.00008 &0.00004 \\
\hline
Solver &Adam &Adam \\
\hline
Momentum and Decay $\beta_1$, $\beta_2$ &0.9, 0.999 &0.9, 0.999 \\
\hline
Decay &$0.0005$ &$0.0005$ \\
\hline
Step &[$150$k, $250$k, $360$k] &[$100$k, $200$k, $250$k] \\
\hline
Step Size &0.5 &0.5 \\
\hline
Total Epochs &400k &300k \\
\hline
\end{tabular}}
\end{center}
\vspace{-5pt}
\caption{Training Parameters for both Image Mode and Video Mode}
\vspace{-5pt}
\label{table:params}
\end{table}

Next, we proceed training in the \textbf{Video Mode} where we expose the network to video sequences. For static image datasets including COCO and MPII, we augment data with video motion sequences by synthesizing motion with scaling, rotation and translation over the unroll count. We train COCO, MPII and PoseTrack in Video Mode with a batch distribution of of $0.4$, $0.1$ and $0.5$, respectively. Moreover, we also use skip-frame augmentation for video-based PoseTrack dataset, where some of the randomly selected sequences skip up to $3$ frames. We lock the weights of VGG module in Video Mode and only train the STAFs and keypoints blocks. For Model I, we only trained the TAFs block when training on videos. For Model II, we trained PAFs, keypoints and TAFs for $5000$ epochs, then locked PAFs and keypoints before training TAFs only. In Model III, both STAFs and keypoints were kept unlocked and were trained for $300$k iterations. 

\subsection{Inference and Tracking} \label{subsec:inference}
The method described till now predicts heatmaps of keypoints and STAFs at every frame by running CNNs associated with each module while passing required data computed from previous frame. Next, we present the framework to perform pose inference as well as tracking across frames given the output heatmaps. Let the inferred poses at time $t$ and $t-1$ be given by:
\begin{align}
\mathbf{P}^{t} = & \;\; \{\mathbf{P}^{t,1}, \mathbf{P}^{t,2}, \ldots, \mathbf{P}^{t,N}\}, \nonumber\\
\mathbf{P}^{t-1} = & \;\; \{\mathbf{P}^{t-1,1}, \mathbf{P}^{t-1,2}, \ldots, \mathbf{P}^{t-1,M}\},
\label{eq:all_poses}
\end{align}
where the second superscript indexes over people in each frame. Each pose at a particular time $\mathbf{P}^{t,n}$ consists of up to $K$ keypoints post inference, i.e., $\mathbf{P}^{t,n}$ includes only those keypoints that become part of a pose:
\begin{align}
\mathbf{P}^{t,n} = & \;\; \{\overline{\mathbf{K}}_1^{t,n}, \overline{\mathbf{K}}_2^{t,n}, \ldots, \overline{\mathbf{K}}_K^{t,n} \}.
\label{eq:all_poses}
\end{align}

The detection and tracking procedure begins with localization of keypoints at time $t$. The inferred keypoints $\overline{\mathbf{K}}^{t}$ are obtained by rescaling the heatmaps to match the original image resolution followed by non-maximal suppression. Then, we infer PAF and TAF weights between all possible pairs of keypoints in each frame defined by the given topology, i.e., 
\begin{equation}
\mathbf{\overline{L}}_{k \rightarrow k'}^{t} 
\omega\big(\mathbf{\overline{K}}_{k}^{t}, \mathbf{\overline{K}}_{k'}^{t}\big),\;\;
\mathbf{\overline{R}}_{k \rightarrow k'}^{t} 
\label{eq:defPAF_TAF}
\end{equation}
where the function $\omega(\cdot)$ samples points between the two argument keypoints, computes the dot product between directional vector among the two points and the mean vector of the sampled points. The inference of PAF, $\overline{\mathbf{L}}^{t}$, and TAF, $\overline{\mathbf{R}}^{t}$, weights is constrained by the spatio-temporal topology, where the spatial and temporal constraints are encoded in tables $\mathbf{\ddot{L}}$ and $\mathbf{\ddot{R}}$, respectively. 

\begin{equation}
max\left(\sum_{t}^{N}\sum_{k}^{K}\left(\overline{\mathbf{L}}_{k\rightarrow k'}^{t,n}+\mathbf{\overline{R}}_{k\rightarrow k'}^{t}\right)\right)
\label{eq:xdef666}
\end{equation} 

Overall, we wish to generate a set of people $\mathbf{P}^{t}$ with ids that maximizes the connection scores between their keypoints pairs $(\overline{\mathbf{K}}_{k}^{t-1,n},\overline{\mathbf{K}}_{k'}^{t,n})$, and temporal connections given previous keypoints $\overline{\mathbf{K}}^{t-1,n}$ from tracklets $\mathbf{P}^{t-1,n}$ given an association.

\begin{algorithm}[t]
\caption{: Estimation and tracking of keypoints and STAFs}
\label{alg:poseInference}

\textbf{Input}: $\mathbf{K}^{t},\mathbf{L}^{t},\mathbf{R}^{t}$, and $\mathbf{P}^{t-1}$ with unique ids \\
\textbf{Output}: $\mathbf{P}^{t}$ with ids

\rule{1\linewidth}{0.02cm}

\begin{algorithmic}[1]
\Procedure{inferPoses}{$ $}
    \State Compute $\overline{\mathbf{L}}^{t}$ given $\overline{\mathbf{K}}^{t}$, $\mathbf{L}^{t}$ and $\mathbf{\ddot{L}}$
    \State Compute $\overline{\mathbf{R}}^{t}$ given $\overline{\mathbf{K}}^{t}$, $\mathbf{R}^{t}$, $\mathbf{P}^{t-1}$ and $\mathbf{\ddot{R}}$
    \State Sort $\overline{\mathbf{L}}^{t}$ and $\overline{\mathbf{R}}^{t}$ by score
    \State Initialize empty map of people $\mathbf{P}^{t}$
	\For{every $\overline{\mathbf{L}}_{k \rightarrow k'}^{t,n}$ in $\overline{\mathbf{L}}^{t}$}
	    \State If $k$ and $k'$ unassigned; add new $\mathbf{P}^{t,n}$
	    \State If $k$ or $k'$ assigned; add to existing $\mathbf{P}^{t,n}$
	    \State If $k$ and $k'$ assigned; update score of $\mathbf{P}^{t,n}$
	    \State If $k$ assigned to $\mathbf{P}^{t,n}$ and $k'$ assigned to $\mathbf{P}^{t,n'}$ \State \& $\mathbf{P}^{t,n}$ and $\mathbf{P}^{t,n'}$ lack opposing points;
	    \State merge $\mathbf{P}^{t,n}$ and $\mathbf{P}^{t,n'}$.
	\EndFor
	\For{$\mathbf{P}^{t,n}$ in $\mathbf{P}^{t}$}
	\For{$\overline{\mathbf{K}}^{t,n}$ in $\mathbf{P}^{t,n}$}
	    \State Find $\overline{\mathbf{R}}^{t}$ with the highest score; copy id
	\EndFor
	\State Update $\mathbf{P}^{t,n}$ with the most frequent id
	\EndFor
	\State If insufficient keypoint matches for $\mathbf{P}^{t,n}$; 
	\State initialize tracklet
	\State Remove $\mathbf{P}^{t-1,n}$ if no association made
\EndProcedure
\end{algorithmic}
\end{algorithm}

To do this, both the inferred PAF and TAF weights are computed then sorted by their scores before inferring the complete poses, $\mathbf{P}^{t}$, and associating them across frames with unique ids. We perform this in a bottom-up style as described in Algorithm~\ref{alg:poseInference} where we utilize $\overline{\mathbf{R}}^{t}$ and $\mathbf{P}^{t-1}$ to determine the update, addition or deletion of tracklets. Going through each PAF in the sorted list, (i) we initialize a new pose if both keypoints in the PAF are unassigned, (ii) add to existing pose if one of the keypoints is assigned, (iii) update score of PAF in pose if both are assigned to the same pose, and (iv) merge two poses if keypoints belong to different poses with opposing keypoints unassigned. Finally, we assign id to each pose in the current frame with the most frequent id of keypoints from the previous frame. This is done over all tracklets and people very quickly as it is done on the GPU.

Furthermore, we make use of past poses $\mathbf{P}^{t-1}$ and TAFs $\overline{\mathbf{R}}^{t}$ to reweigh PAFs. For cases where we have an ambiguous PAFs (Alg. \ref{alg:poseInference}, 7:) as seen in Figure~\ref{fig:transitivity}, we use transitivity that reweighs PAFs to disambiguate between them. In this figure, keypoint $\{A\}$ - an elbow - is under consideration, with wrists $\{B\}$/$\{E\}$ as two possibilities. We select the strongest TAFs where $\{A,B,C,D,A\}$ has a higher weight than $\{A,E,F,G,A\}$.

\begin{equation}
\overline{\mathbf{L}}_{k \rightarrow k'}^{t,n} = (1-\alpha)\omega(\overline{\mathbf{K}}_{k}^{t-1,n},\overline{\mathbf{K}}_{k'}^{t,n})
+ \alpha*\omega(\overline{\mathbf{K}}_{k}^{t,n},\overline{\mathbf{K}}_{k'}^{t,n}).
\label{eq:xdef55}
\end{equation}


\section{Additional Experiments}
\label{sec:experiments}
We present some experiments that were otherwise not displayed in detail in the main paper.

For the sake of completion, we first report results on the COCO dataset in Table~\ref{table:coco}. Despite using single set of weights for all the stages, we were able to get close results. Our network is designed to be lightweight and work in a recurrent fashion, so our main reference point is still the Posetrack datasets.

\begin{table}[h]
\footnotesize
\begin{center}
\resizebox{0.475\textwidth}{!}{
\setlength{\tabcolsep}{6pt}
\begin{tabular}{l |c| c c c c }
\hline
\textbf{Method} & \textbf{AP} & \textbf{AP$^{50}$} & \textbf{AP$^{75}$} & \textbf{AP$^{M}$}	& \textbf{AP$^{L}$} \\
\hline
\multicolumn{6}{ c }{Top-Down Approaches} \\
Megvii~\cite{chen2017cascaded}	&73.0&	91.7&	80.9&	69.5	&78.1	\\
G-RMI~\cite{papandreou2017towards}&	71.0&	87.9&	77.7&	69.0&	75.2	\\
Mask R-CNN~\cite{he2017maskrcnn}&	69.2&	90.4&	76.0&	64.9&	76.3\\
\hline
\multicolumn{6}{ c }{Bottom-Up Approaches} \\
PersonLab~\cite{papandreou2018personlab} & 68.7 & 89.0 & 75.4 & 64.1 & 75.5 \\
Associative Emb.~\cite{newell2017} & 65.5 & 86.8 & 72.3 & 60.6 & 72.6\\
OpenPose 2018~\cite{gines} & 64.4 & 86.5 & 70.2 & 61.5 & 68.8 \\
\hline
Proposed & 61.5 & 82.2 & 67.1 & 58.5 & 66.7 \\
\hline
\end{tabular}}
\end{center}
\caption{Results on the COCO test-dev dataset. Top: top-down results. Bottom: bottom-up results (top methods only). AP$^{50}$ is for OKS $=0.5$, AP$^{L}$ is for large scale persons.}
\label{table:coco}
\end{table}

\smallskip
\noindent\textbf{Filter Sizes:} We observed that having each \cross{7} filter replaced with a three \cross{3} filter resulted in better accuracies, especially for knees and ankles. The results are shown in Table~\ref{table:3x3vs7x7}. We run single frame inference on Model I and find the \cross{3} to be $2\%$ more accurate than \cross{7}, with significant boosts in average precision of knee and ankle keypoints.

\begin{figure}[h]
\centering
\includegraphics[width=0.25\textwidth]{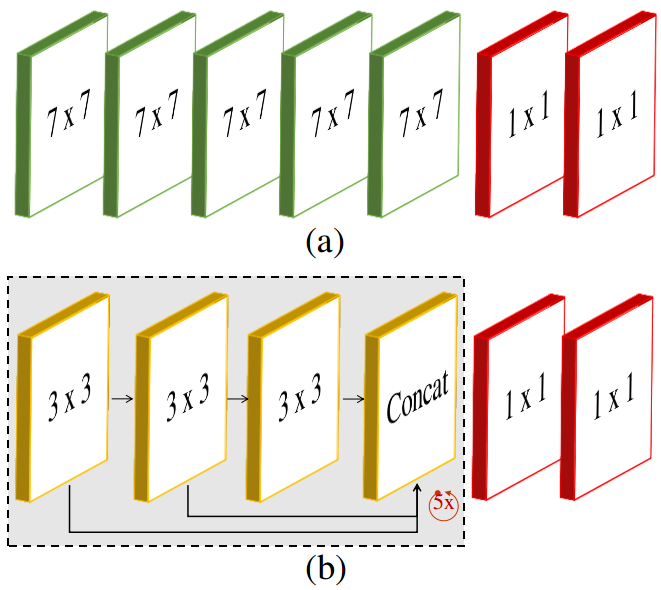}
\caption{Types of filters usued}
\label{fig:ghosting}
\end{figure}
 
\begin{table}[h]
\begin{center}
\resizebox{0.475\textwidth}{!}{
\setlength{\tabcolsep}{4pt}
\begin{tabular}{l | c c c c c c c | c | c}
\hline
\textbf{Method} & \textbf{Hea} & \textbf{Sho} & \textbf{Elb} & \textbf{Wri} & \textbf{Hip} & \textbf{Kne} & \textbf{Ank} & \textbf{mAP} & \textbf{fps} \\
\hline
Model I - 3x3 & 75.7 &73.9 &67.8 &56.3 &66.8 &62.3 &56.9 &66.3 &14\\
\hline
Model I - 7x7 & 76.0 &73.3 &66.4 &54.0 &63.4 &59.2 &52.2 &64.3 &10\\
\hline
\end{tabular}}
\end{center}
\vspace{-5pt}
\caption{This table shows results for experiments with the two filter sizes on PoseTrack 2017 validation set.}
\vspace{-5pt}
\label{table:3x3vs7x7}
\end{table}

\smallskip
\noindent\textbf{Recurrence of Keypoints Module:} To verify that the network was indeed benefiting from ingesting previous frame heatmaps, we explicitly train a model where the keypoint module was connected to current PAF module in an \textit{auxiliary} fashion and did not receive previous heatmaps (first row of Table~\ref{table:keypoint_auxvscon}). The second row shows the case where keypoint module was \textit{connected} to ingest output heatmaps from previous frames. The result is $2.1\%$ improvement at single scale on the PoseTrack 2017 validation set using Model II with \cross{7} filter size.

\begin{table}[h]
\begin{center}
\resizebox{0.475\textwidth}{!}{
\setlength{\tabcolsep}{4pt}
\begin{tabular}{l | c c c c c c c | c | c}
\hline
\textbf{Method} & \textbf{Hea} & \textbf{Sho} & \textbf{Elb} & \textbf{Wri} & \textbf{Hip} & \textbf{Kne} & \textbf{Ank} & \textbf{mAP} & \textbf{fps} \\
\hline
KP Auxiliary & 72.3 &71.2 &63.9 &51.4 &60.1 &56.3 &50.0 &61.5 &28\\
\hline
KP Connected & 76.2 &71.6 &64.5 &51.9 &62.6 &59.3 &52.5 &63.6 &27\\
\hline
\end{tabular}}
\end{center}
\vspace{-5pt}
\caption{Performance using Model II where the keypoint module does not take feedback from previous heatmaps (auxiliary), and when it does ingest previous heatmaps (connected).}
\vspace{-5pt}
\label{table:keypoint_auxvscon}
\end{table}

\noindent\textbf{STAF Topology:} We experimented with Topology A, B and C. Topology B proved to be better than A due to it's ability to preserve information even during minimal motion, lending itself better to the recurrent structure of our network. It especially performed better during jittery camera motion, or during crowded scenes with several people. Topology C, which does not consist of any current frame spatial information, was difficult to train and resulted in an MAP that was about 8\% lower. This was mainly because we had to construct a person during the first frame, or during a new person appearance, and simply propagate it using the TAF, which proved to be less reliable than extracting poses on every frame with PAF/TAF, then propagating it with TAF. 

\section{Implementation}

Training: We train on 4 Titan XP in Caffe. Testing: We test on a single 1080 Ti, and i7 7800X. We write our own code in C++ and CUDA.

\FloatBarrier
{\small
\bibliographystyle{ieee}
\bibliography{egbib}
}

\end{document}